\pdfoutput=1
\documentclass[11pt]{article}
\usepackage{booktabs}
\usepackage{acl2023}

\usepackage{overpic}
\usepackage{color,soul}

\usepackage{times}
\usepackage{dashrule}
\usepackage{arydshln}
\usepackage{latexsym}
\usepackage{multirow}
\usepackage{subcaption}
\usepackage{amsfonts,amssymb}
\usepackage{graphicx}
\usepackage{CJK}
\usepackage{makecell}
\usepackage[utf8]{inputenc}
\usepackage{microtype}
\usepackage[T1]{fontenc}
\usepackage[utf8]{inputenc}
\usepackage{microtype}
\usepackage{amsmath}
\usepackage{color}
\usepackage{colortbl}
\usepackage{multirow}
\usepackage{verbatim}
\usepackage{stfloats}
\newcommand{\name}{E-NER}

\title{\name: Evidential Deep Learning for Trustworthy Named Entity Recognition}

\author{Zhen Zhang\textsuperscript{1} \quad Mengting Hu\textsuperscript{1}\thanks{\; Mengting Hu and Bingzhe Wu are the corresponding authors.} \quad 
Shiwan Zhao\textsuperscript{}\thanks{\; Independent researcher.} \quad 
Minlie Huang\textsuperscript{2} \quad 
Haotian Wang\textsuperscript{1} \\ 
\textbf{Lemao Liu\textsuperscript{3} \quad 
Zhirui Zhang\textsuperscript{3} \quad 
Zhe Liu\textsuperscript{4} \quad 
Bingzhe Wu\textsuperscript{3}}\textsuperscript{*} \\
\textsuperscript{1} College of Software, Nankai University,
\textsuperscript{2} The CoAI group, Tsinghua University \\
\textsuperscript{3} Tencent AI Lab, \textsuperscript{4} Zhejiang Lab\\
\tt zhangz@mail.nankai.edu.cn, \tt mthu@nankai.edu.cn}

\begin{document}
\maketitle
\begin{abstract}

Most named entity recognition (NER) systems focus on improving model performance, ignoring the need to quantify model uncertainty, which is critical to the reliability of NER systems in open environments. Evidential deep learning (EDL) has recently been proposed as a promising solution to explicitly model predictive uncertainty for classification tasks. However, directly applying EDL to NER applications faces two challenges, i.e., the problems of \textit{sparse entities} and \textit{OOV/OOD entities} in NER tasks. To address these challenges, we propose a trustworthy NER framework named {\name~}\footnote{\href{https://github.com/Leon-bit-9527/ENER}{{https://github.com/Leon-bit-9527/ENER}}} by introducing two uncertainty-guided loss terms to the conventional EDL, along with a series of uncertainty-guided training strategies. Experiments show that \name~ can be applied to multiple NER paradigms to obtain accurate uncertainty estimation. Furthermore, compared to state-of-the-art baselines, the proposed method achieves a better OOV/OOD detection performance and better generalization ability on OOV entities. 
\end{abstract}

\section{Introduction}\label{Introduction}
\noindent Named entity recognition (NER) aims to locate and classify entities in unstructured text, such as extracting LOCATION information \textit{"New York"} from the sentence \textit{"How far is New York from me"}. 
Thanks to the development of deep neural network (DNN), 
current NER methods have achieved remarkable performance on a wide range of benchmarks \cite{lample,yamada-etal-2020-luke,word2word}. 

\begin{figure}
    \centering
    \includegraphics[scale=0.67]{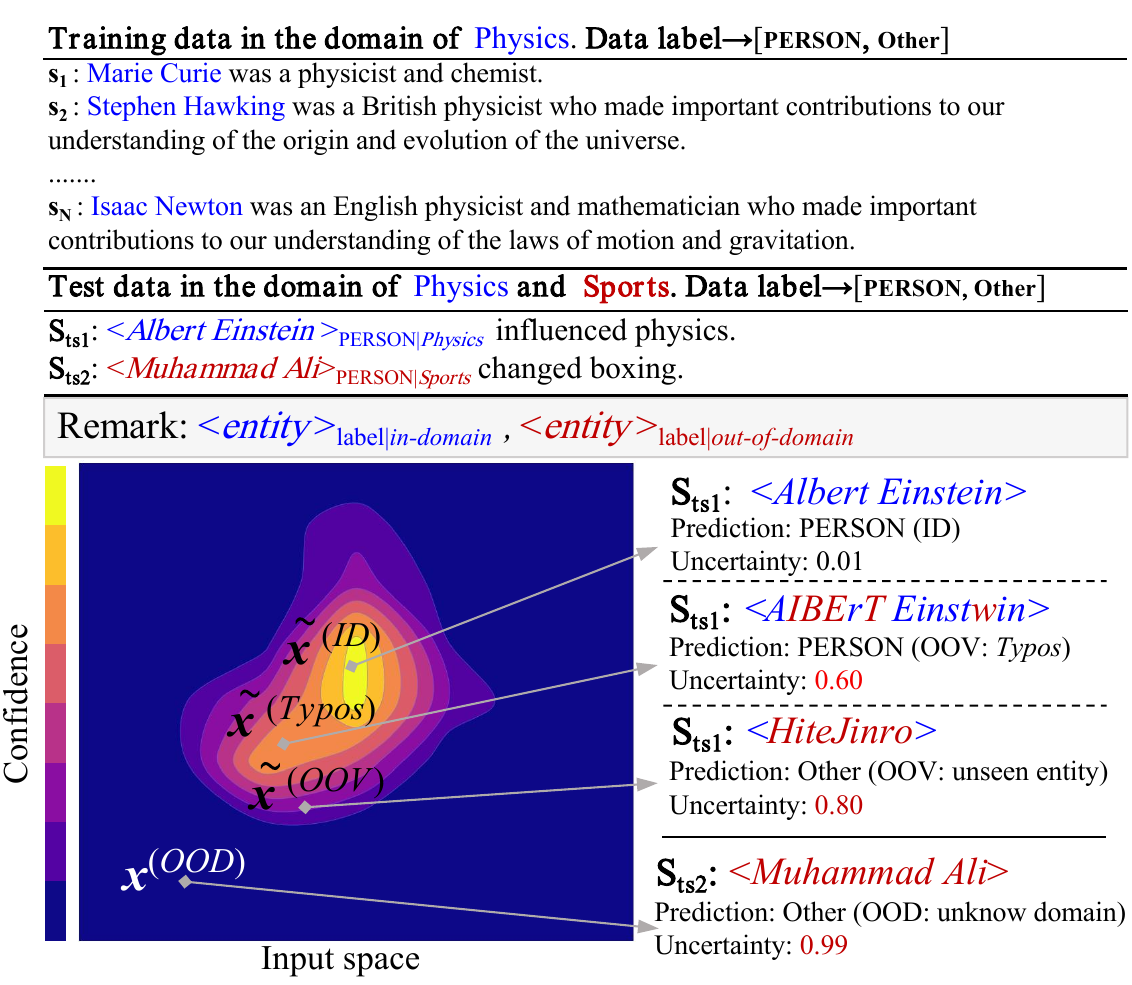}
    \caption{Visualization of desired uncertainty estimations in the NER application.}
    \label{fig:motivation}
\end{figure}

Despite this progress, current NER-related research typically focuses on improving the model performance, such as recognition accuracy and F1 scores \cite{yu-etal-2020-named, boundary}. However, seldom works focus on investigating the model's reliability. The critical aspect of the model reliability is the uncertainty estimation of the predictive results, which can characterize the probability that the model prediction will be wrong. One natural way to construct the predictive uncertainty is based on the maximum value of the Softmax output \cite{seq2seq,word2word,boundary} (the smaller this value, the larger the uncertainty). However, previous empirical studies show that probabilistic predictions produced by DNN models (e.g., transformer and CNN) are often inaccurate \cite{Calibration,Confidence_calibrated_OOD,pinto2022impartial}. Therefore, this natural way may over/under-estimate the predictive uncertainty, hindering the model's reliability. 

High-quality uncertainty estimation helps to improve the model's reliability in an open environment and to find valuable samples to improve training sample efficiency, thus reducing the cost of manual labeling. On the one hand, for the reliability aspect, accurate uncertainty estimation can equip the NER model with the ability to express \textit{``I do not know''} to both the out-of-domain (OOD) or out-of-vocabulary (OOV) samples \cite{posterior_network}. A desired uncertainty estimation is conceptually shown in Figure \ref{fig:motivation}, wherein misclassified OOV/OOD entities are assigned with significantly higher uncertainty than the in-domain (ID) entities. Besides, the estimated uncertainty can be further absorbed into the training process to improve the model robustness against OOV/OOD samples. On the other hand, for the sample efficiency aspect, prior work shows that high-quality uncertainty estimation can also be used for selecting more "informative" samples and thus can reduce the number of labeled samples required for training the NER model.

To attain high-quality uncertainty estimation, evidential deep learning (EDL) \cite{MuratSensoy2018EvidentialDL} provides a promising solution.
EDL is superior to existing Bayesian learning-based methods \cite{blundell2015weight,kingma2015variational, graves2011practical} in that model
uncertainty can be efficiently estimated in a single forward pass that avoids inexact posterior approximation \cite{dbu} or time/storage-consuming Monte Carlo sampling \cite{gal2016dropout}.
However, directly applying conventional EDL to NER applications still faces two critical challenges: (1) \textit{sparse entities}: In text corpus, entities only take a minority. For example, only 16.8\% of the words in the commonly used CoNLL2003 dataset belong to entities. The remaining non-entity types are labeled into the "others" (O) class. The imbalance between entity and non-entity words can cause over-fitting and poor performance on the entity types. (2) \textit{OOV/OOD entity discrimination}: In the open environment, NER training/test data typically comes with OOV/OOD entities. However, the optimization objective of current EDL methods lacks explicit modeling of such types of information. 

To address these two issues, we present a trustworthy NER framework named \name~ with a series of uncertainty-guided training strategies. For the issue of sparse entities, we propose to use an uncertainty-guided importance weighted (IW) loss, wherein samples with higher predictive uncertainties are assigned larger weights. This loss helps the model training to pay more attention to entities of interest (e.g., person and location). To solve the issue of unknown entities, we present an additional regularization term to penalize the case where labels are more prone to errors by assigning higher uncertainties to corresponding samples. We empirically show these two uncertainty-guided loss terms can improve both the quality of estimated confidence and the robustness against OOV samples.

Our contributions are summarized as follows:
\begin{itemize}
    \item To the best of our knowledge, \name~ is the first work to explore how to leverage evidential deep learning to improve the reliability of current NER models. This work has successfully shown the potential of EDL to provide high-quality uncertainty estimation in NER applications. The estimated uncertainty can be further used for detecting OOD/OOV samples in the test phase.
    \item For the technique contribution, we propose two uncertainty-guided loss terms to mitigate sparse entities and OOV/OOD entity discrimination issues in the NER task.  
    \item \name~ is extensively validated in a series of experiments. In contrast to conventional NER methods, the result shows that \name~ comes with the following superiority: (1) more accurate uncertainty estimation. (2) better OOV/OOD detection performance. (3) better generalization ability on OOV entities. (4) better sample efficiency (i.e., fewer samples are required to achieve the same-level performance).
    
\end{itemize}
    
\begin{figure*}[t]
    \centering
    \includegraphics[scale=0.53]{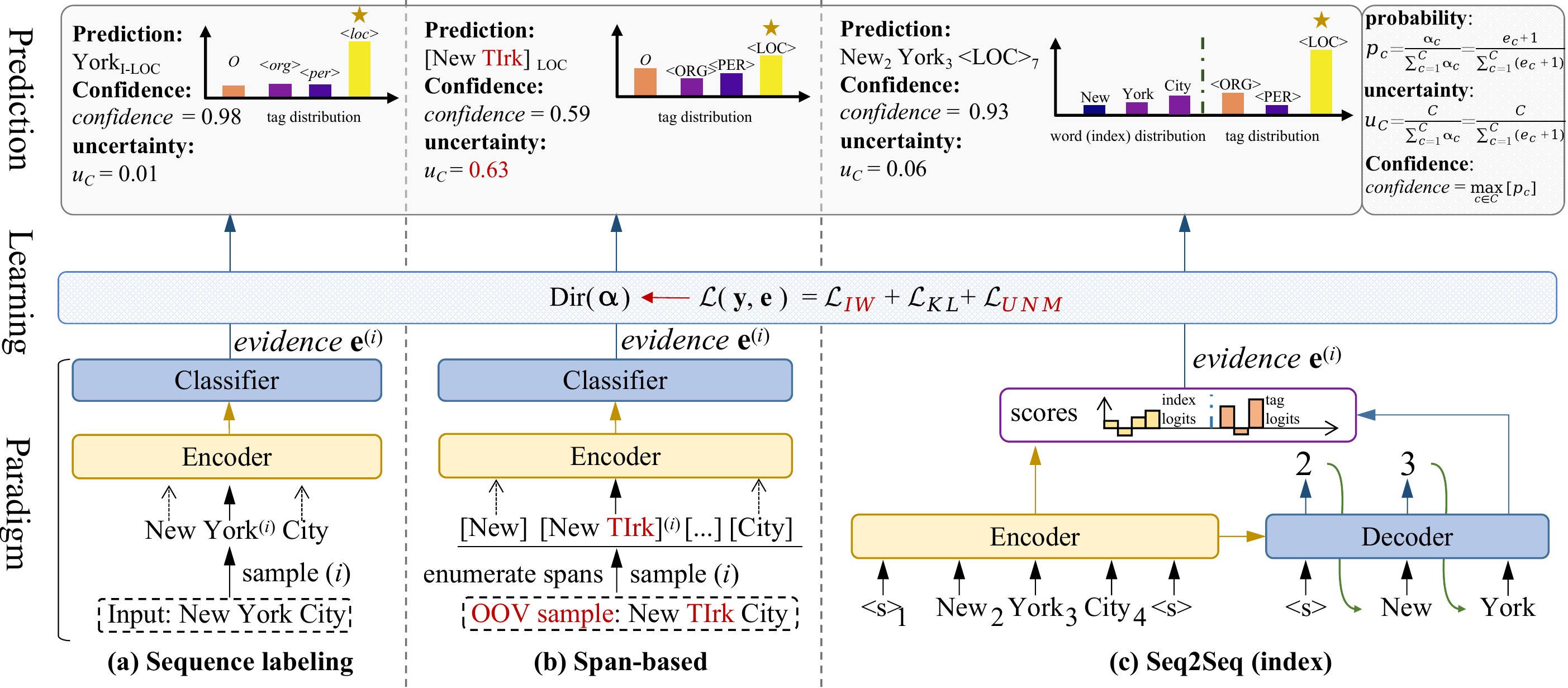}
    \caption{ Basic architecture of \name~with different NER paradigms.}
    \label{fig:framework}
\end{figure*}
\section{Preliminary}\label{Preliminary}
\noindent 
This section introduces a commonly-used EDL implementation based on the Dirichlet-based model (DBM) \cite{MuratSensoy2018EvidentialDL}. We then describe how the DBM computes the uncertainty in a closed form. 
\subsection{Dirichlet-based Model}\label{preliminary_dbm}
Conventional neural network classifiers typically employ a Softmax layer to provide a point estimation of the categorical distribution. In contrast, Dirichlet-based models (DBM) output the parameters of a Dirichlet distribution and then use it to estimate the categorical distribution. Specifically, for the $i$-th sample $x^{(i)}$ (e.g., the $i$-th word in the NER task) in the $C$-class classification task, the DBM replaces the Softmax of the neural network with an activation function layer (e.g., Softplus) to ensure that the network outputs non-negative values, which are considered as the evidence $\mathbf{e}^{(i)}\in \mathbb{R}^{C}_{+}$ to support the classification. The evidence is then used for constructing a Dirichlet distribution which models the distribution over different classes. To this end, the parameter of a Dirichlet distribution is obtained by: $\boldsymbol{\alpha}^{(i)} = \mathbf{e}^{(i)} + \mathbf{1}$, where $\mathbf{1}$ represents the vector of $C$ ones. Finally, the density function of Dirichlet distribution is given by:
\begin{equation}
    \begin{aligned}
        {\rm{Dir}}(\mathbf{p}^{(i)}|\boldsymbol{\alpha}^{(i)})=\frac{1}{B(\boldsymbol{\alpha}^{(i)})}\prod_{c=1}^{C}p_{c}^{(\alpha^{(i)}_{c}-1)},
    \end{aligned}\label{dir_density}
\end{equation}
where $B(\boldsymbol{\alpha}^{(i)})$ is the $C$-dimensional multinomial beta function.

To learn model parameters, given the sample $(x^{(i)},\mathbf{y}^{(i)})$, where $\mathbf{y}^{(i)}$ is a one-hot $C$-dimensional label for sample $x^{(i)}$, previous EDL methods build the optimization objective by combining a cross-entropy classification loss $\mathcal{L}_{CLS}$ and a KL penalty loss $\mathcal{L}_{KL}$:
\begin{equation}
\begin{aligned}
\mathcal{L}^{(i)}_{EDL}\!=&\mathcal{L}^{(i)}_{CLS}+\mathcal{L}^{(i)}_{KL}\\
    =&\underbrace{\sum_{c=1}^C y^{(i)}_{c}\left(\psi(S^{(i)}) -\psi(\alpha^{(i)}_{c})\right)}_{\text{(a) classification loss}}\\+& 
    \underbrace{\lambda_1 KL[{\rm{Dir}}(\mathbf{p}^{(i)}|{\widetilde{\boldsymbol{\alpha}}^{(i)})}||{\rm{Dir}}(\mathbf{p}^{(i)}|\mathbf{1})]}_{\text{(b) penalty loss}},
    \label{eq:edl}
\end{aligned}
\end{equation}
where $\psi(\cdot)$ is the digamma function, and $S^{(i)}\!=\!\sum^{C}_{c=1}\alpha^{(i)}_{c}$ denotes the Dirichlet strength, $\lambda_1 $ is the balance factor, ${\rm{Dir}}(\mathbf{p}^{(i)}|\mathbf{1})$ is a special case which is equivalent to the uniform distribution, and $\widetilde{\boldsymbol{\alpha}}^{(i)} = \mathbf{y}^{(i)} + (1-\mathbf{y}^{(i)}) \odot \boldsymbol{\alpha}^{(i)}$ denotes the masked parameters while $\odot$ refers to the Hadamard (element-wise) product, which removes the non-misleading evidence from predicted parameters $\boldsymbol{\alpha}^{(i)}$. Intuitively, the first term in Eq. \ref{eq:edl} measures the classification performance while the second term can be seen as a regularization term that penalizes misleading evidences by encouraging the associate distribution to be close to uniform distribution (see more details in Appendix \S \ref{DBM_Optimization}).

\subsection{Uncertainty Estimation of DBM}
Once we obtain the Dirichlet distribution for prediction, we can estimate the predictive uncertainty in a closed form. To this end, EDL provides two probabilities: \textit{belief mass} and \textit{uncertainty mass}. %These probabilities will be used later in the training process. 
The belief mass $\mathbf{b}$ represents the probability of evidence assigned to each category and the uncertainty mass $u$ %quantifies the level of uncertainty and 
provides uncertainty estimation. Specifically, for the sample $x^{(i)}$, %in $c^{th}$ class
the belief mass $b^{(i)}_c$ and uncertainty $u^{(i)}$ are computed as:   
\begin{equation}
    b^{(i)}_c = \frac{{e}^{(i)}_{c}}{S^{(i)}}\quad\text{and}\quad u^{(i)} =\frac{C}{S^{(i)}},
    \label{eq:b}
\end{equation}
with the restrictions that $u^{(i)} + \sum^{C}_{c=1} b^{(i)}_{c}= 1$. The belief mass $\mathbf{b}$ and the uncertainty mass $u$ will be used to guide the training process in our proposed framework (see Section \S\ref{uncer_learning}). 
\section{\name~Architecture}\label{Architecture}
\noindent In this section, we describe the three core modules of E-NER and provide an overview of the system architecture in Figure \ref{fig:framework}. Additionally, we revise the learning strategy of EDL by incorporating importance weights (IW) to address the sparse entities problem and uncertainty mass optimization (UNM) to model the uncertainty of mispredicted entities.

\subsection{NER Feature Extraction}
\noindent Given a word sequence $X = \{x^{(1)},...,x^{(n)}\}$ and a target sequence $Y = \{y^{(1)},...,y^{(n)}\}$. To obtain the hidden representation $H$ of $X$, the words in the sentence $X$ are first preprocessed according to the input form required by the corresponding NER method. Then the processed input is fed into an Encoder module (e.g., BERT \cite{bert}) to compute the hidden representation $H= {\rm{Encoder}}(X)$, where $H\in \mathbb{R}^{n \times d_h}$ and $d_h$ denotes the dimension of the hidden representation. The input format for NER models can vary depending on the paradigm used. Three NER paradigms were considered for this study: sequence labeling (Figure \ref{fig:framework}(a)), span-based (Figure \ref{fig:framework}(b)), and Seq2Seq (Figure \ref{fig:framework}(c)). The specific formats for these paradigms are provided in the Appendix \S \ref{NER_paradigms}. Note that in the Seq2Seq (sequence-to-sequence) paradigm, we choose a pointer-based model \cite{seq2seq}, so that we don't need to learn on the entire vocabulary.

\subsection{Dirichlet-based Prediction Layer}
\noindent Once we obtain the hidden representation, we introduce a Dirichlet-based layer to produce the final predictive distribution. Precisely, for the $i^{th}$ sample, the hidden representation $h$ is fed to the fully connected layer to output logits, and then we can transform the logits into Dirichlet parameters $\boldsymbol{\alpha}$ as described in Section \S\ref{preliminary_dbm}.
Finally, as shown in Figure \ref{fig:framework}, only one forward step using Eq. \ref{eq:b} is sufficient to calculate the uncertainty $u^{(i)}$, while the probability distribution $\mathbf{ p}^{(i)}$ and prediction ${y}^{(i )}$ are calculated as follows:
\begin{equation}
    {\mathbf{p}^{(i)}}\!=\!\frac{\boldsymbol{\alpha}^{(i)}}{S^{(i)}}, \; \quad y^{(i)}\!=\!\mathop{\mathrm{arg\,max}}\limits_{c\in C}\left[{{p}^{(i)}_{c}}\right].
    \label{eq:p_predict}
\end{equation}
\subsection{\name~Model Learning}\label{uncer_learning}
\noindent \textbf{Overview}. The objective function of EDL training is to minimize the sum of losses over all words. Due to the \textit{sparse entities} and \textit{OOV/OOD entities} issues, directly applying EDL to NER leads %to overfitting to non-entities, which in turn leads to 
suboptimal uncertainty estimates. We improve conventional EDL methods by incorporating belief mass and uncertainty into the network training process. Specifically, two key modifications are introduced: (1) We compute importance weights for each sample based on the belief mass to reweight the original classification loss in Eq. \ref{eq:edl}(a). (2) We introduce an additional term to increase the uncertainty of mispredicted instances, which explicitly improves the quality of uncertainty estimation and helps OOD entity detection.
\begin{figure}[t]
    \centering
    \includegraphics[scale=0.8]{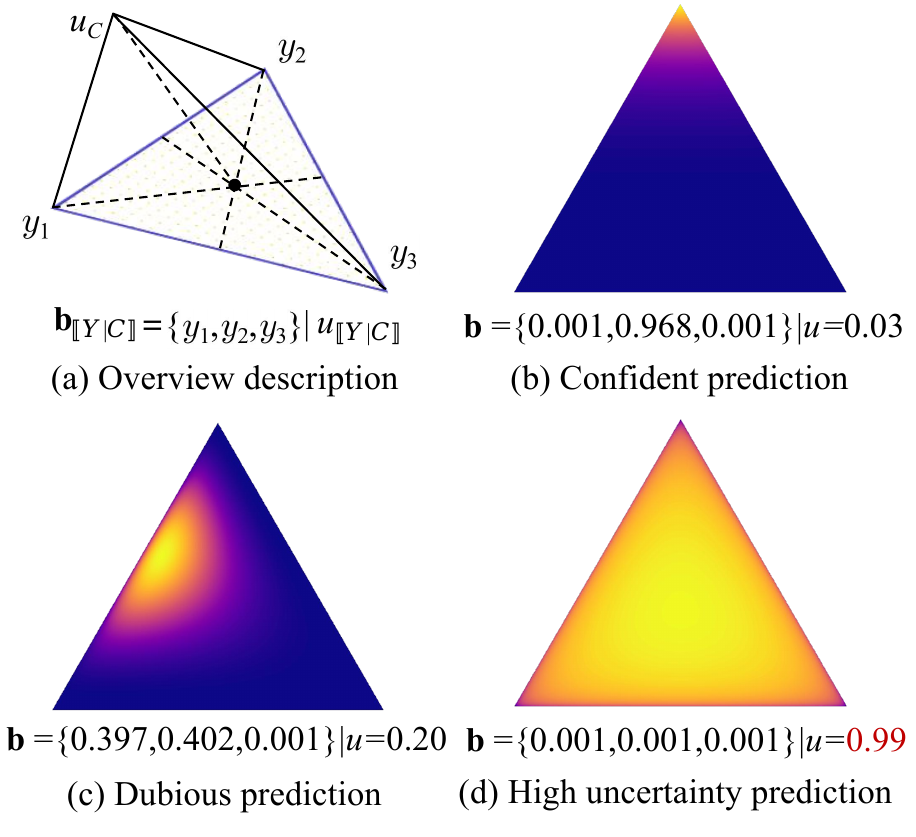}
    \caption{(a) Overview of uncertainty estimation for Dirichlet distributions. (b-d) Typical patterns of Dirichlet distribution for an example 3-class classification task.}
    \label{fig:diri_case}
\end{figure}

\noindent \textbf{Importance Weight}.
Due to the inherent imbalance between entities and non-entities in NER datasets, conventional EDL methods tend to overfit non-entities and assign high uncertainty estimates to entities. To make the training focus more on the entities and increase the evidence corresponding to the ground-truth category, we use the belief mass of the ground-truth category to compute the category-level uncertainty for each instance to adjust the loss. Specifically, for the $i^{th}$ sample, we use $(\mathbf{1}-\mathbf{b}^{(i)})$ as the category-level uncertainty which serves as the importance weights of entity categories during training. To this end, we replace the ground truth $\mathbf{y}^{(i)}$ of one-hot representation with an importance weight (IW) $\mathbf{w}^{(i)}=(\mathbf{1}-\mathbf{b}^{(i)})\odot \mathbf{y}^{(i)}$, and lastly, the Eq. \ref{eq:edl}(a) is adjusted to:
\begin{equation}
\begin{aligned}
    \mathcal{L}^{(i)}_{IW}=& \sum_{c=1}^C \textcolor[rgb]{0.80,0.10,0.1}{w^{(i)}_{c}}\left(\psi(S^{(i)}) -\psi(\alpha^{(i)}_{c})\right).
    \label{eq:IW}
\end{aligned}
\end{equation}

As illustrated in Figure \ref{fig:diri_case}(b), the belief mass of the ground-truth category is high, indicating a high level of certainty in the prediction. In this case, the importance weight (IW) assigned will be small. Conversely, Figure \ref{fig:diri_case}(c) presents a small belief mass, indicating an uncertain prediction. IW will be assigned a large value. In this manner, the training process can focus more on sparse but valuable entities.

\noindent \textbf{Uncertainty Mass Optimization}. 
Assigning high uncertainty to OOV/OOD entities (see Figure \ref{fig:diri_case}(d) as an example) facilitates OOV/OOD entity detection. However, ground-truth OOV/OOD samples are not available during training. One solution is to synthesize such data on the boundary of the in-domain region via a generative model~\cite{Confidence_calibrated_OOD}. In this paper, we propose a more convenient way to treat hard samples as OOV/OOD samples which are often outliers and are mispredicted even after adequate model training. In this way, we enable the model to detect OOV/OOD data. Specifically, uncertainty mass optimization (UNM) assigns higher uncertainty to more error-prone samples for the model to express a lack of evidence, by adding an uncertainty mass penalty term $\mathcal{L}_{UNM}$ to the wrongly predicted samples:
\begin{equation}
    \begin{aligned}
        \mathcal{L}_{UNM} \!=\!-\!\lambda_2\!\sum_{i \in \{ \hat {y}^{(i)} \ne y^{(i)}\}}\!{\rm{log}}(u^{(i)}).
    \end{aligned}
    \label{eq:l3}
\end{equation}
The coefficient $\lambda_2 = \lambda_0 \ \mathrm {exp}\{-(\mathrm{ln}\lambda_0/T)t\}$, where $\lambda_2 \in [\lambda_0,1]$, $\lambda_0 \ll 1$ is a small positive constant, $t$ is the current training epoch, and $T$ is the total number of training epochs.  As the training epoch $t$ increases towards $T$, the factor $\lambda_2$ will increase monotonically from $\lambda_0$ to 1.0. This allows the network to initially focus on optimizing classification and gradually shift its emphasis towards optimizing UNM as the training progresses.

\noindent \textbf{Overall Loss}. The overall loss function combines three components: the importance weighted classification loss $\mathcal{L}_{IW}$, the KL divergence penalty loss $\mathcal{L}_{KL}$, and the uncertainty mass loss $\mathcal{L}_{UNM}$ for mispredicted entities. Each element contributes to the overall loss and is defined as follows:
\begin{equation}
    \mathcal{L}_{overall}=\sum_{i=1}^{N}(\mathcal{L}^{(i)}_{IW} + 
    \mathcal{L}^{(i)}_{KL}) + \mathcal{L}_{UNM}.
    \label{eq:l2}
\end{equation}

\section{Experiments}\label{Experiments}

\subsection{Research Questions}
In this section, we design extensive experiments to validate whether the proposed method obtains high-quality uncertainty estimation. Concretely, the following four research questions will be investigated.

\textbf{RQ1:} Whether \name~ improves the quality of confidence estimation in contrast to prior work?

\textbf{RQ2:} Can uncertainty provided by E-NER achieve better OOV/OOD detection performance?

\textbf{RQ3:} Can \name~ improve the model generalization ability on OOV samples? 

\textbf{RQ4:} Can \name~ help to find valuable instances to improve the sample efficiency of NER model training?

Following these four research questions, we provide further discussions on our method including ablation studies and limitations.

\begin{table}\small\centering
    \begin{tabular}{lllll}
        \toprule  %添加表格头部粗线
        Dataset& Sentences& Types& Domain  \\
        \midrule  %添加表格中横线
        CoNLL2003& \multicolumn{1}{c}{22,137}& \multicolumn{1}{c}{4}& \multicolumn{1}{c}{Newswire}\\
        OntoNotes 5.0& \multicolumn{1}{c}{76,714}& \multicolumn{1}{c}{18}&\multicolumn{1}{c}{General}\\
        WikiGold& \multicolumn{1}{c}{1,696}& \multicolumn{1}{c}{4}&\multicolumn{1}{c}{General}\\
        \bottomrule %添加表格底部粗线
    \end{tabular}
    \caption{Statistics of the NER dataset.}
    \label{tab:dataset}
\end{table}

\begin{table}\small\centering
    \begin{tabular}{lllll}
        \toprule  %添加表格头部粗线
        Dataset& Sentences& Entities&  OOV Rate  \\
        \midrule  %添加表格中横线
        TwitterNER& \multicolumn{1}{c}{3257}& \multicolumn{1}{c}{3990}& \multicolumn{1}{c}{0.62}\\
        CoNLL2003-Typos& \multicolumn{1}{c}{2676}&  \multicolumn{1}{c}{4130}&\multicolumn{1}{c}{0.71}\\
        CoNLL2003-OOV& \multicolumn{1}{c}{3685}& \multicolumn{1}{c}{5648}&\multicolumn{1}{c}{0.96}\\
        \bottomrule %添加表格底部粗线
    \end{tabular}
    \caption{Statistics of OOV entities in the test set.}
    \label{tab:OOVData}
\end{table}

\subsection{Datasets and Metrics}
\textbf{Datasets from Different Domains}. To answer the above research questions, we choose three widely-used datasets, including CoNLL2003 \cite{CONLL}, OntoNotes 5.0 \cite{OntoNotes_dataset}\footnote{\href{https://catalog.ldc.upenn.edu/LDC2013T19}{{https://catalog.ldc.upenn.edu/LDC2013T19}}} and WikiGold \cite{wikigold}. The statistics are displayed in Table \ref{tab:dataset}. 

%\vspace{4pt}
\noindent
\textbf{OOV Datasets}. We further choose three public OOV datasets, including TwitterNER \cite{zhang2018adaptive}, CoNLL2003-Typos \cite{typos}, and CoNLL2003-OOV \cite{typos}. The statistics are displayed in Table \ref{tab:OOVData}.

\noindent\textbf{Metrics}. We evaluate the results using three metrics: F1, Expected Calibration Error (ECE), and Area Under the ROC Curve (AUC). F1 is a commonly used performance indicator in NER. ECE is a metric that measures the confidence calibration of a model, with a low score indicating a well-calibrated model. AUC is a commonly used metric for evaluating the performance of binary classifiers, and we use it to evaluate the OOV/OOD detection performance. Their detailed computations are described in the Appendix \S \ref{implement_details_metrics}. 

\begin{table*}[t]\small
    \begin{center}
    \begin{tabular}{llllllll}
        \toprule  %添加表格头部粗线
            \multirow{2}*{Setting}& \multicolumn{2}{c}{Typos} & \multicolumn{2}{c}{OOV}& \multicolumn{2}{c}{OOD}\\
        &Con&Unc&Con&Unc&Con&Unc\\
        \midrule
        \multicolumn{1}{l}{BERT-Tagger \cite{bert}}&0.812&0.812&0.689&0.751&0.674&0.756&\\
        \quad -EDL&0.805&0.808&0.699&0.759&0.693&0.767&\\
        \quad -\name(ours)&\textbf{0.820}&\textbf{0.817}&\textbf{0.700}&\textbf{0.760}&\textbf{0.769}&\textbf{0.799}&\\
        \cdashline{1-8}
        \multicolumn{1}{l}{SpanNER\cite{fu-etal-2021-spanner}}&0.717&0.783&0.614&0.773&0.623&0.799& \\
        \quad -EDL&0.701&0.759&0.607&0.760&0.620&0.792&\\
        \quad -\name(ours)&\textbf{0.741}&\textbf{0.792}&\textbf{0.640}&\textbf{0.796}&\textbf{0.676}&\textbf{0.824}&\\
        \cdashline{1-8}
        \multicolumn{1}{l}{Seq2Seq \cite{seq2seq}}&0.825&0.833&0.724&0.794&0.797&0.820\\
        \quad -EDL&\textbf{0.829}&0.830&0.729&0.787&0.793&0.818 \\
        \quad -\name(ours)&0.824&\textbf{0.841}&\textbf{0.743}&\textbf{0.803}&\textbf{0.822}&\textbf{0.847}\\
        \bottomrule %添加表格底部粗线
    \end{tabular}
    \end{center}
    \caption{Evaluation results of OOV/OOD detection in terms of AUC. The three binary detection tasks can use either confidence (Con) or uncertainty (Unc) for classification.}
    \label{tab:rq2}
\end{table*}

\subsection{Experiment Setting}

We conduct experiments on three popular NER paradigms: sequence labeling, span-based, and Seq2Seq. The following three models are chosen for evaluating each paradigm. 

\noindent
\textbf{BERT-Tagger} \cite{bert}. It follows the classical paradigm, recognizing entities via \emph{sequence labeling}.

\noindent
\textbf{SpanNER}\footnote{\href{https://github.com/neulab/spanner}{{https://github.com/neulab/spanner}}}
\cite{fu-etal-2021-spanner}. It enumerates all spans and detects entities from them. For simplicity, we use the original span-based method, without any constraints or data processing.

\noindent
\textbf{Seq2Seq}\footnote{\href{https://github.com/yhcc/BARTNER}{{https://github.com/yhcc/BARTNER}}}  \cite{seq2seq}. It is a generative model based on BART, which does not require additional labeling strategies and entity enumeration.

In the experiments, all the reported results are the average of five runs. The experiment details are introduced in Appendix \S \ref{implement_details}.

\begin{table}[t]\small
    \begin{center}
    \begin{tabular}{lllll}
        \toprule  %添加表格头部粗线
        % \cmidrule[0.8pt]{1-5}
            \multirow{2}*{Setting}& \multicolumn{2}{c}{CoNLL2003} & \multicolumn{2}{c}{OntoNotes 5.0}\\
        & F1($\uparrow$)& ECE($\downarrow$)& F1($\uparrow$)& ECE($\downarrow$)\\
        \cmidrule[0.8pt]{1-5}
        \multicolumn{1}{l}{BERT-Tagger}& 91.32&0.0845&  88.20& 0.1053  \\
        \quad -EDL& 91.36&0.0755& 88.09& 0.0838 \\
        \quad -\name(ours)& \textbf{91.55}&\textbf{0.0739}& \textbf{88.74}& \textbf{0.0603} \\
        \cdashline{1-5}
        \multicolumn{1}{l}{SpanNER}& 91.94& 0.0673& 87.82& 0.0609  \\
        \quad -EDL& 91.97&0.0481&  87.39& 0.0474 \\
        \quad -\name(ours)& \textbf{92.06}&\textbf{0.0414}& \textbf{88.44}& \textbf{0.0434}  \\
        \cdashline{1-5}
        \multicolumn{1}{l}{Seq2Seq}& 93.05& 0.0324 & 89.89 & 0.0375 \\
        \quad -EDL& 92.84& 0.0322& 90.22&0.0329 \\
        \quad -\name(ours)& \textbf{93.15}& \textbf{0.0225}& \textbf{90.64}&\textbf{0.0328} \\
        \bottomrule %添加表格底部粗线
        % \cmidrule[0.8pt]{1-5}
    \end{tabular}
    \end{center}
    \caption{Evaluation results in various NER systems, in terms of F1 (\%) and ECE for evaluating performance and confidence quality, respectively.}
    \label{tab:confidence_estimation}
\end{table}

\begin{figure}[t]
\centering
\includegraphics[width=0.43\textwidth]{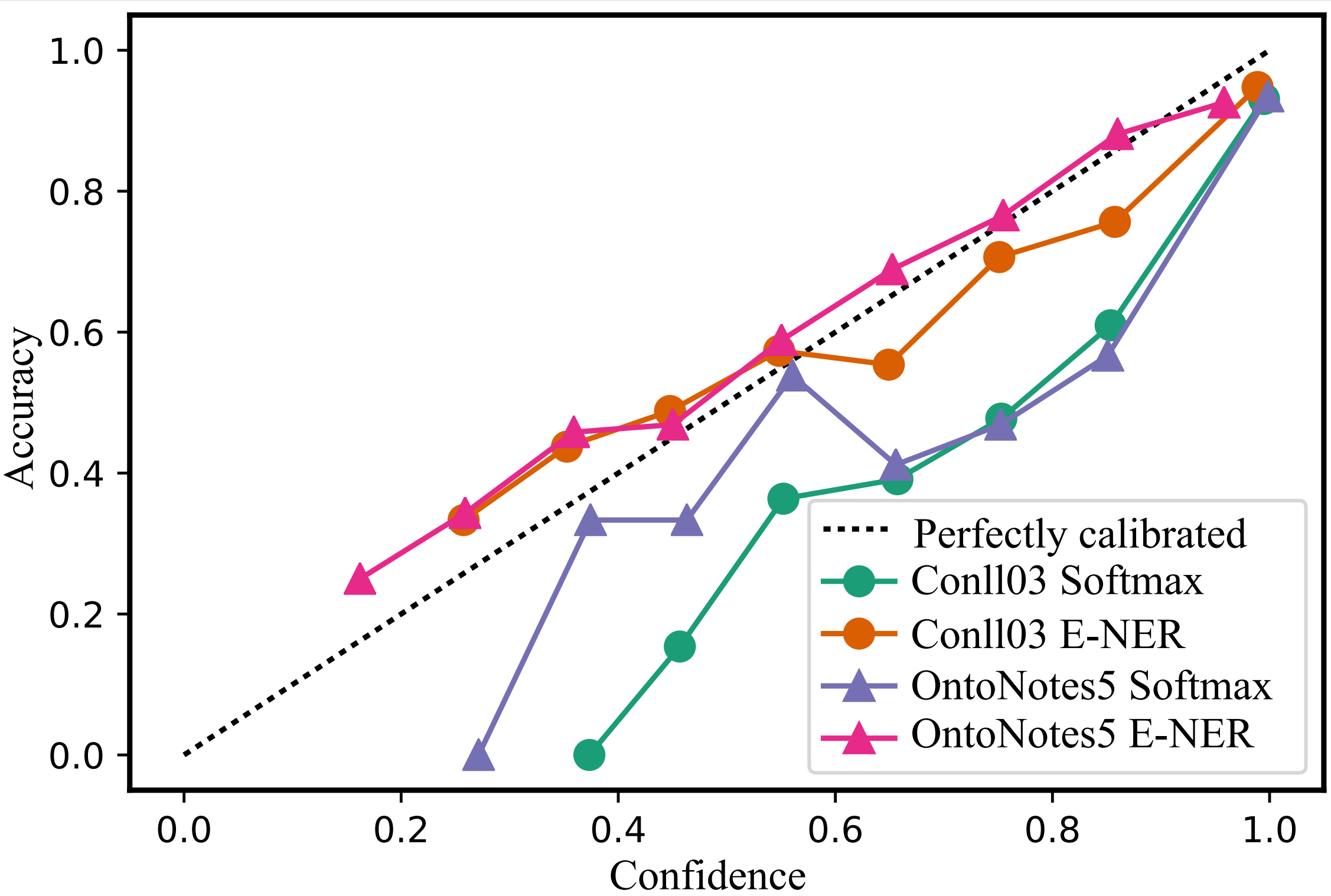}
    \caption{Model calibration curve. The basic encoder is SpanNER. This figure is depicted by evaluating subsets separately, where each subset has the same range of confidence.}
    \label{fig:mocel_calibration} %图片名称和图片标号
\end{figure}

\subsection{Research Question Discussions}
\subsubsection{Confidence Estimation Quality}\label{rq1}
To answer the first research question, an important concept should be clarified, i.e., \emph{what is qualified confidence?} This concept should have a positive correlation with performance, meaning that higher confidence should indicate better performance and vice versa, as depicted by the dashed line in Figure \ref{fig:mocel_calibration}. Our findings reveal that on both datasets, Softmax is far below the perfectly calibrated line, indicating that confidence does not reflect performance well, and it is an example of \emph{over-confidence}. However, E-NER is found to approach the perfect calibrated line. This suggests that E-NER can produce well-qualified confidence.

We further evaluate all paradigms and present the results in Table \ref{tab:confidence_estimation}. It can be observed that E-NER consistently performs the best across all paradigms. This demonstrates that E-NER can be effectively applied in various frameworks. When comparing EDL to the original models, it is observed that while EDL improves confidence estimation, it also results in a decline in performance. For example, on OntoNotes 5.0 dataset, EDL performs worse than BERT-Tagger and SpanNER in terms of the F1 metric. This highlights the limitations of directly applying the EDL approach. In contrast, E-NER performs the best on both metrics, demonstrating that it can provide better-qualified confidence without negatively impacting performance, and even achieving slight improvements in all settings. A typical reliability diagram is also included in Appendix \S \ref{relia_diagrams} for a more detailed representation. 

\subsubsection{OOV/OOD Detection}
The typical usage of uncertainty is to detect whether an instance is OOV/OOD or not, as large uncertainty tends to reveal unnatural instances, such as OOV and OOD.  To evaluate uncertainty from this usage (RQ2), we choose three binary detection tasks, including typos, OOV, and OOD. The results are shown in Table \ref{tab:rq2}.

Firstly, it can be observed that, when compared to the original model of each paradigm, EDL does not improve the performances in most experiments of the three paradigms. This verifies that EDL is not effective in addressing the \emph{OOV/OOD entity discrimination} challenge of NER. Then we found that E-NER significantly outperforms the original models and EDL in various paradigms. In particular, in span-based OOD detection, E-NER outperforms SpanNER by +5.3\% and EDL by +5.6\% on AUC when using confidence for detection. This demonstrates the effectiveness of E-NER in distinguishing whether an entity is OOV/OOD or not. Note that using uncertainty is better than using confidence for OOV/OOD detection in most cases. 

\linespread{1.1}
\begin{table}[t]\small
    \begin{center}
    \setlength\tabcolsep{3pt}
    \begin{tabular}{p{2cm}p{1.5cm}p{1cm}p{1cm}p{1cm}p{1cm}}
        \toprule  %添加表格头部粗线
            \multirow{2}*{Methods}& \multirow{2}*{TwitterNER}&
            \multicolumn{2}{c}{CoNLL2003}\\
            \cline{3-4}
            & &\multicolumn{1}{c}{Typos}&\multicolumn{1}{c}{OOV}\\
        \midrule  %添加表格中横线
        \multicolumn{1}{l}{VaniIB \cite{AlemiFD016}}&
        \multicolumn{1}{c}{71.19}&
        \multicolumn{1}{c}{83.49}&
        \multicolumn{1}{c}{\textbf{70.12}}\\
        \multicolumn{1}{l}{DataAug \cite{DA}}&
        \multicolumn{1}{c}{73.69}
        &\multicolumn{1}{c}{81.73}
        &\multicolumn{1}{c}{69.60}\\
        \multicolumn{1}{l}{SpanNER (BERT large)}&
        \multicolumn{1}{c}{71.57}&
        \multicolumn{1}{c}{81.83}&
        \multicolumn{1}{c}{64.43}\\
        \multicolumn{1}{l}{SpanNER (RoBERTa large)}&
        \multicolumn{1}{c}{71.70}&
        \multicolumn{1}{c}{82.85}&
        \multicolumn{1}{c}{64.70}\\
        \multicolumn{1}{l}{SpanNER (AlBERT large)}&
        \multicolumn{1}{c}{70.33}&
        \multicolumn{1}{c}{82.49}
        &\multicolumn{1}{c}{64.12}\\
        \cdashline{1-4}
        \multicolumn{1}{l}{EDL-SpanNER (BERT large)}&\multicolumn{1}{c}{74.14}
        &\multicolumn{1}{c}{82.89}
        &\multicolumn{1}{c}{68.40}\\
        \multicolumn{1}{l}{E-SpanNER (BERT base)}&
        \multicolumn{1}{c}{74.94}
        &\multicolumn{1}{c}{83.31}&
        \multicolumn{1}{c}{67.99}\\
        \multicolumn{1}{l}{E-SpanNER (BERT large)}&\multicolumn{1}{c}{\textbf{75.64}}
        &\multicolumn{1}{c}{\textbf{83.64}}
        &\multicolumn{1}{c}{69.71}\\
        \cdashline{1-4}
        \multicolumn{1}{l}{$\Delta$ \name-NER vs. SpanNER}&\multicolumn{1}{c}{\textcolor[rgb]{0,0,1}{4.07↑}}
        &\multicolumn{1}{c}{\textcolor[rgb]{0,0,1}{1.81↑}}
        &\multicolumn{1}{c}{\textcolor[rgb]{0,0,1}{5.28↑}}\\
        \bottomrule %添加表格底部粗线
    \end{tabular}
    \end{center}
    \caption{Evaluation results of generalization on OOV samples in terms of F1 (\%). To compare fairly, we also choose SpanNER as the basic encoder.}
    \label{tab:oov_results}
\end{table}

\subsubsection{Generalization on OOV Samples}
Another benefit of well-qualified confidence is the robustness to noise, since the model is properly calibrated without over or under-confidence. Thus, we further investigate E-NER's generalizing ability on OOV samples (RQ3). The results on three OOV datasets are reported in Table \ref{tab:oov_results}.

It is first observed that E-NER (BERT large) achieves the best performances on TwitterNER and CoNLL2003-Typos datasets, and competitive performance on CoNLL2003-OOV. Compared with a strong baseline SpanNER (BERT large), E-NER (BERT large) significantly outperforms it by +4.07\%, +1.81\% and +5.28\% on three datasets, respectively. This validates the generalizing ability of our approach. Secondly, by comparing EDL (BERT large) and E-NER s(BERT large), our method also achieves consistently better performances. This further validates that our proposed two uncertainty-guided loss terms effectively promote the robustness against OOV samples.

\begin{table}[t]\small
    \begin{center}
    \begin{tabular}{llllll}
        \toprule  %添加表格头部粗线
            \multirow{2}*{Setting}& \multicolumn{2}{c}{CoNLL2003}&\multicolumn{2}{c}{OntoNotes 5.0}\\
            &\multicolumn{1}{c}{Ratio}&\multicolumn{1}{c}{F1($\uparrow$)}&\multicolumn{1}{c}{Ratio}&\multicolumn{1}{c}{F1($\uparrow$)}\\
        \midrule  %添加表格中横线
        \multicolumn{1}{l}{Random} &5.5\%&85.39&3.0\%&79.47\\
        \multicolumn{1}{l}{Entropy}&5.5\%&88.29&3.0\%&84.80\\
        \multicolumn{1}{l}{MC dropout}&5.5\%&88.67&3.0\%&86.06\\
        \multicolumn{1}{l}{EDL}&5.5\%&90.51&3.0\%&86.25\\
        \multicolumn{1}{l}{\name} &5.5\%&\textbf{90.88}&3.0\%&\textbf{86.68}\\
        % \multicolumn{1}{c}{\Delta} &\multicolumn{1}{c}{-} &\textcolor[rgb]{0,0,1}{5.49↑}
        % &\multicolumn{1}{c}{-}
        % &\textcolor[rgb]{0,0,1}{7.21↑}\\
        \bottomrule %添加表格底部粗线
    \end{tabular}
    \end{center}
    \caption{Evluation results of in-domain data selection in terms of F1 (\%). Ratio indicates the proportion of selected samples out of the whole training set.}
    \label{tab:rq4_1}
\end{table}

\begin{table}[t]\small
    \begin{center}
    \begin{tabular}{llllll}
        \toprule  %添加表格头部粗线
            \multirow{2}*{Setting}& \multicolumn{2}{c}{${\rm{WikiGold}}_{\gets{\rm{CoNLL.}}}$}&\multicolumn{2}{c}{${\rm{CoNLL2003}}_{\gets{\rm{Onto.}}}$}\\
            &\multicolumn{1}{c}{Ratio}
            &\multicolumn{1}{c}{F1($\uparrow$)}
            &\multicolumn{1}{c}{Ratio}
            &\multicolumn{1}{c}{F1($\uparrow$)}\\
        \midrule  %添加表格中横线
        \multicolumn{1}{l}{Random} &4.8\%&\multicolumn{1}{c}{53.67}&4.7\%&\multicolumn{1}{c}{84.23}\\
        \multicolumn{1}{l}{Entropy} &4.8\%&\multicolumn{1}{c}{80.63}&4.7\%&\multicolumn{1}{c}{88.81}\\
        \multicolumn{1}{l}{MC dropout} &4.8\%&\multicolumn{1}{c}{82.87}&4.7\%&\multicolumn{1}{c}{90.32}\\
        \multicolumn{1}{l}{EDL}&4.8\%&\multicolumn{1}{c}{83.32}&4.7\%&\multicolumn{1}{c}{90.12}\\
        \multicolumn{1}{l}{\name} &4.8\%&\multicolumn{1}{c}{\textbf{84.08}}&\multicolumn{1}{c}{4.7\%}&\multicolumn{1}{c}{\textbf{90.52}}\\
        % \multicolumn{1}{c}{\Delta} &\multicolumn{1}{c}{-} &\multicolumn{1}{c}{\textcolor[rgb]{0,0,1}{30.50↑}}
        % &\multicolumn{1}{c}{-}
        % &\multicolumn{1}{c}{\textcolor[rgb]{0,0,1}{6.29↑}}\\
        \bottomrule %添加表格底部粗线
    \end{tabular}
    \end{center}
    \caption{Evaluation results of cross-domain data selection in terms of F1 (\%). The left side of the arrow $\gets$ is the target domain, and the right side is the source domain.}
    \label{tab:rq4_2}
\end{table}

\subsubsection{Sample Efficiency}
In active learning, a sample's uncertainty can be utilized for data selection. Then whether the selected samples are valuable also suggests the quality of uncertainty. To 
evaluate E-NER from this perspective (RQ4), we design in-domain and cross-domain sample selection experiments. The results are displayed in Table \ref{tab:rq4_1} and Table \ref{tab:rq4_2}, respectively.  

It is found that using the same scale of samples, E-NER achieves consistently the best performances in both the in-domain and cross-domain settings. This verifies that uncertainty predicted by E-NER has better quality. Concretely, MC dropout attains uncertainty with multiple runs of sub-models, which costs time and memory. Though outperforming naive random selection and entropy of softmax, MC dropout is still less performed than EDL and E-NER, which both directly compute the uncertainty in one forward pass. Then we see that EDL does not always outperform MC dropout, as the cross-domain experiment ${\rm{CoNLL2003}}_{\gets{\rm{Onto}}}$ shown. Yet E-NER, concentrating on two issues of NER task, is universally effective, and can better handle the challenges of an open environment.

\subsection{Further Analysis}

\begin{table}\small
    \begin{center}
    \setlength\tabcolsep{4pt}
    \begin{tabular}{llllll}
        \toprule  %添加表格头部粗线
        % \cmidrule[0.8pt]{1-13}
            \multirow{2}*{Setting}& \multicolumn{2}{c}{CoNLL2003} & \multicolumn{2}{c}{OntoNotes 5.0}\\
        &\multicolumn{1}{c}{F1}& \multicolumn{1}{c}{ECE}& \multicolumn{1}{c}{F1}&\multicolumn{1}{c}{ECE} \\
        \midrule
        %{Softmax}&91.94&0.067&87.82&0.061&85.96&0.082&\\
        %EDL&91.97&0.048&87.39&0.047&85.52&0.075&\\
        \name&92.06&\textbf{0.041}&\textbf{88.44}&\textbf{0.043}\\ 
        ~-UNM&\textbf{92.10}&0.058&88.21&0.051\\
        ~-IW&91.95&0.045&87.77&0.042 \\
        
        % \cmidrule[0.8pt]{1-13}
        \bottomrule
    \end{tabular}
    \end{center}
    \caption{Evaluation results of ablation study in terms of F1 (\%) and ECE.}
    \label{tab:ablation}
\end{table}

\textbf{Ablation Study}. To explore the effects of individual loss terms, the ablation study is presented in Table \ref{tab:ablation}. It is observed that removing each loss term would cause performance declines in most evaluation metrics. Concretely, removing IW causes the F1 score to decrease more than removing UNM. On the contrary, removing UNM makes a significant degradation in ECE. Overall, this study indicates that the proposed uncertainty-guided terms are both effective.

\noindent
\textbf{Why E-NER Works}. We incorporate two uncertainty-guided loss terms into EDL. Firstly, IW is designed for sparse entities which leads to an imbalance problem. Using uncertainties as weights helps the model training to pay more attention to entities of interest. As reported in Table \ref{tab:ablation}, IW is effective in improving the F1 score. Secondly, UNM is proposed to deal with OOV/OOD entities. Such entities should have larger uncertainties compared to normal ones, however, naive EDL does not model this explicitly. E-NER increases the uncertainty of mispredictions which are relatively close to OOV/OOD entities. As shown in Table \ref{tab:ablation}, UNM helps to improve the quality of uncertainty estimation. These two uncertainty-guided loss terms target different NER issues, and using uncertainty (IW) and learning uncertainty (UNM) interactively allows E-NER to perform well in various experimental settings. Furthermore, we showcase actual predictions in Appendix \S \ref{error_case_stduy}.
  
\section{Related Work}\label{{RelatedWork}}

\noindent\textbf{NER Paradigm}. NER is a fundamental task in information extraction. The mainstream methods of NER can be divided into three categories: sequence labeling, span-based, and Seq2Seq. Sequence labeling methods assign a label to each token in a sentence to identify flat entities, and are better at handling longer entities with lower label consistency \cite{fu-etal-2021-spanner}. Span-based methods, which enumerate and classify entity sets in a sentence according to the maximum span length, perform better on sentences with OOV words and entities of medium length \cite{AlemiFD016,DA,fu-etal-2021-spanner}. Seq2Seq methods directly generate the entities and corresponding labels in the sentence, and are capable of handling various NER subtasks uniformly \cite{seq2seq}. Recently, NER systems are undergoing a paradigm shift \cite{akbik-etal-2018-contextual, TENER}, using one paradigm to handle multiple types of NER tasks. \citet{zhang-etal-2022-de} analysis the incorrect bias in Seq2Seq from the perspective of causality, and designed a data augmentation method based on the theory of backdoor adjustment, making Seq2Seq more suitable for unified NER tasks.

\noindent\textbf{Uncertainty Estimation}. Bayesian deep learning uses Bayesian principles to estimate uncertainty in DNN parameters. However, modeling uncertainty in network parameters does not guarantee accurate estimation of predictive uncertainty \cite{sensoy2021misclassification}. Recently, there has been a trend in using the output of neural networks to estimate the parameters of the Dirichlet distribution for uncertainty estimation \cite{MuratSensoy2018EvidentialDL,prior_network}. The EDL \cite{MuratSensoy2018EvidentialDL} has the advantages of generalizability and low computational cost, making it applicable to various tasks \cite{TMC,hu2021uncertainty}. However, their uncertainty estimates have difficulty expressing uncertainties outside the domain \cite{edl_ood1,hu2021uncertainty}. In contrast, the Prior Networks \cite{prior_network} require OOD data during training to distinguish in-distribution (ID) and OOD data. When the NER model encounters unseen entities (e.g., OOV and OOD), it is easy to make unreliable predictions, which are often considered from the perspective of data augmentation or information theory \cite{fukuda-etal-2020-robust, wang-etal-2022-miner}, but there is no guarantee that these methods will achieve a balance between performance and robustness.

\section{Conclusion}\label{Conclusion}

In this work, we study the problem of trustworthy NER by leveraging evidential deep learning. To address the issues of \textit{sparse entities} and \textit{OOV/OOD entities}, we propose E-NER with two uncertainty-guided loss terms. Extensive experimental results demonstrate that the proposed method can be effectively applied to various NER paradigms. The uncertainty estimation quality of E-NER is improved without harming performance. Additionally, the well-qualified uncertainties contribute to detecting OOV/OOD, generalization, and sample selection. These results validate the superiority of E-NER on real-world problems. 

\section*{Limitations}\label{Limitations}
\noindent Our work is the first attempt to explore how evidential deep learning can be used to improve the reliability of current NER models. Despite the improved performance and robustness, our work has limitations that may guide our future work.

First, we propose a simple method to treat hard samples (such as outliers) in the dataset as OOV/OOD samples, enabling the model to detect OOV/OOD data with minimal cost. However, there is still a certain gap between these hard samples and the real OOV/OOD data. OOV/OOD detection performance can still be improved by further incorporating more real OOV/OOD samples, for example, real OOD data from other domains, well-designed adversarial examples, generated OOV samples by data augmentation techniques, etc.

Second, we evaluate the versatility of E-NER by applying it to mainstream NER paradigms. However, there are still other paradigms, such as Hypergraph-based methods \cite{hypergraphs} and the W$^2$NER \cite{word2word} approach in recent work, that could be evaluated in the future.

\section*{Acknowledgements}
We sincerely thank all the anonymous reviewers for providing valuable feedback. This work is supported by the youth program of National Science Fund of Tianjin, China (Grant No. 22JCQNJC01340), the Fundamental Research Funds for the Central University, Nankai University (Grant No. 63221028), and the key program of National Science Fund of Tianjin, China (Grant No. 21JCZDJC00130)

% \bibliography{acl_main}
% \bibliographystyle{acl_natbib}
\clearpage
\appendix
\newcommand{\tabincell}[2]{\begin{tabular}{@{}#1@{}}#2\end{tabular}}  
\begin{table*}[t!]\small
    \centering
    \begin{tabular} {p{50pt}p{100pt}p{115pt}p{135pt}}
    \toprule  %添加表格头部粗线
    \multicolumn{1}{c}{}&\multicolumn{1}{c}{BERT-Tagger}&\multicolumn{1}{c}{SpanNER}&\multicolumn{1}{c}{Seq2Seq}\\
    %\midrule[0.4pt]
    %Boundary&labeling strategy (i.e., BIO)&enumerate the span set&begin- and end- index\\
    \midrule[0.4pt]
    Input&$X=\{x^{(1)},x^{(2)},...,x^{(n)}\}$&$X=\{x^{(1)},x^{(2)},...,x^{(n)}\}$ &$X=\{x^{(1)},x^{(2)},...,x^{(n)}\}$\\

    \midrule[0.4pt]
    Processing & - & \tabincell{l}{Enumerate all spans \\ $S=\{s^{(1)},s^{(2)},...,s^{(m)}\}$} & \tabincell{l}{Obtain start and end indexes of entities \\ $Y=\{y^b_1, y^e_1, y_1, ..., y^b_k, y^e_k, y_k\}$}
    \\
    \midrule[0.4pt]
    Hidden state&\tabincell{l}{$h = {\rm{Encoder}}(X); $\\$h\in \mathbb{R}^{n \times d}$}&\tabincell{l}{$h = {\rm{Encoder}}(s^{(i)}); $\\$h\in \mathbb{R}^{d}$}&\tabincell{l}{$h_{t} = \mathrm{EncoderDecoder}(X, Y_{<t});$ \\ $h_{t}\in \mathbb{R}^{d}$}\\
    \midrule[0.4pt]
    Inference&Token-level classification&Span-level classification&Target sequence $Y$ generation\\
    \bottomrule %添加表格底部粗线
    \end{tabular}
    \caption{Explanation of the three NER paradigms.}
    \label{tab:paradigms}
\end{table*}

\section{NER Paradigms}\label{NER_paradigms}
\noindent Here we introduce three popular NER paradigms, shown in Table \ref{tab:paradigms}.

\vspace{3pt}
\noindent\textbf{BERT-Tagger}. It follows the sequence labeling paradigm, which aims to assign a tagging label $Y = \{y^{(1)},...,y^{(n)}\}$ to each word in a sequence $X = \{x^{(1)},...,x^{(n)}\}$. We use BERT-Tagger 
\cite{bert} as the baseline method for sequence labeling. The labeling method adopts a BIO tag set, which indicates the beginning and interior of an entity, or other words. $X$ is fed to BERT to obtain hidden states, followed by a nonlinear classifier to classify each word.

\vspace{3pt}
\noindent\textbf{SpanNER}. Given an input sentence $X = \{x^1,...,x^n\}$, SpanNER enumerates all spans and obtains a set $S=\{s^{(1)},...,s^{(i)},...,s^{(m)}\}$. Then it assigns each span an entity label ${y}$ \cite{fu-etal-2021-spanner}. The maximum length $l$ of the span is artificially set. Assume a sentence's length is \textit{n} and the maximum span length is set to 2, the subscript of the span set can be expressed as $\{(1,1),(1,2)...(n-1,n-1), (n-1,n),(n,n)\}$. Each span is fed into the encoder to obtain a vector representation.

\vspace{3pt}
\noindent\textbf{Seq2Seq}. As presented in Table \ref{tab:paradigms}, given an input sentence {$X=\{x^{(1)},x^{(2)},...,x^{(n)}\}$, the target sequence is represented as $Y = \{y^b_1, y^e_1, y_1, ...,y^b_k, y^e_k, y_k\}$. This target sequence indicates $X$ describes $k$ entities. Take the first entity as an example, its beginning and end indexes are $y^b_1$ and $y^e_1$, with entity category $y_1$. This method learns in a sequence-to-sequence manner \cite{seq2seq}. 

\begin{figure}[t!]
    \begin{subfigure}[b]{0.22\textwidth}
        \includegraphics[width=\textwidth]{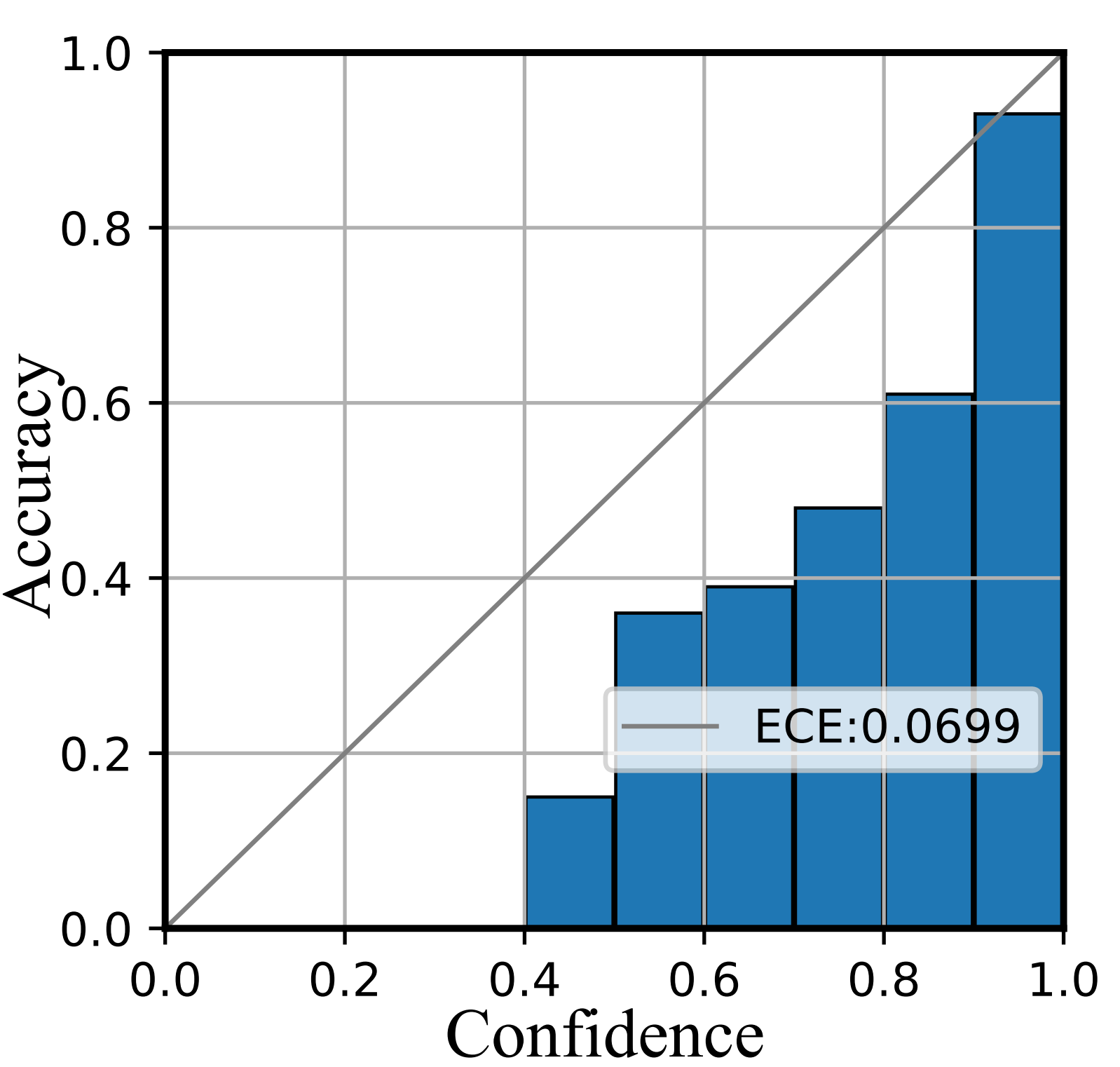}
        \caption{CoNLL2003 Softmax}
        \label{fig:ECE1}
    \end{subfigure}
    \begin{subfigure}[b]{0.22\textwidth}
        \includegraphics[width=\textwidth]{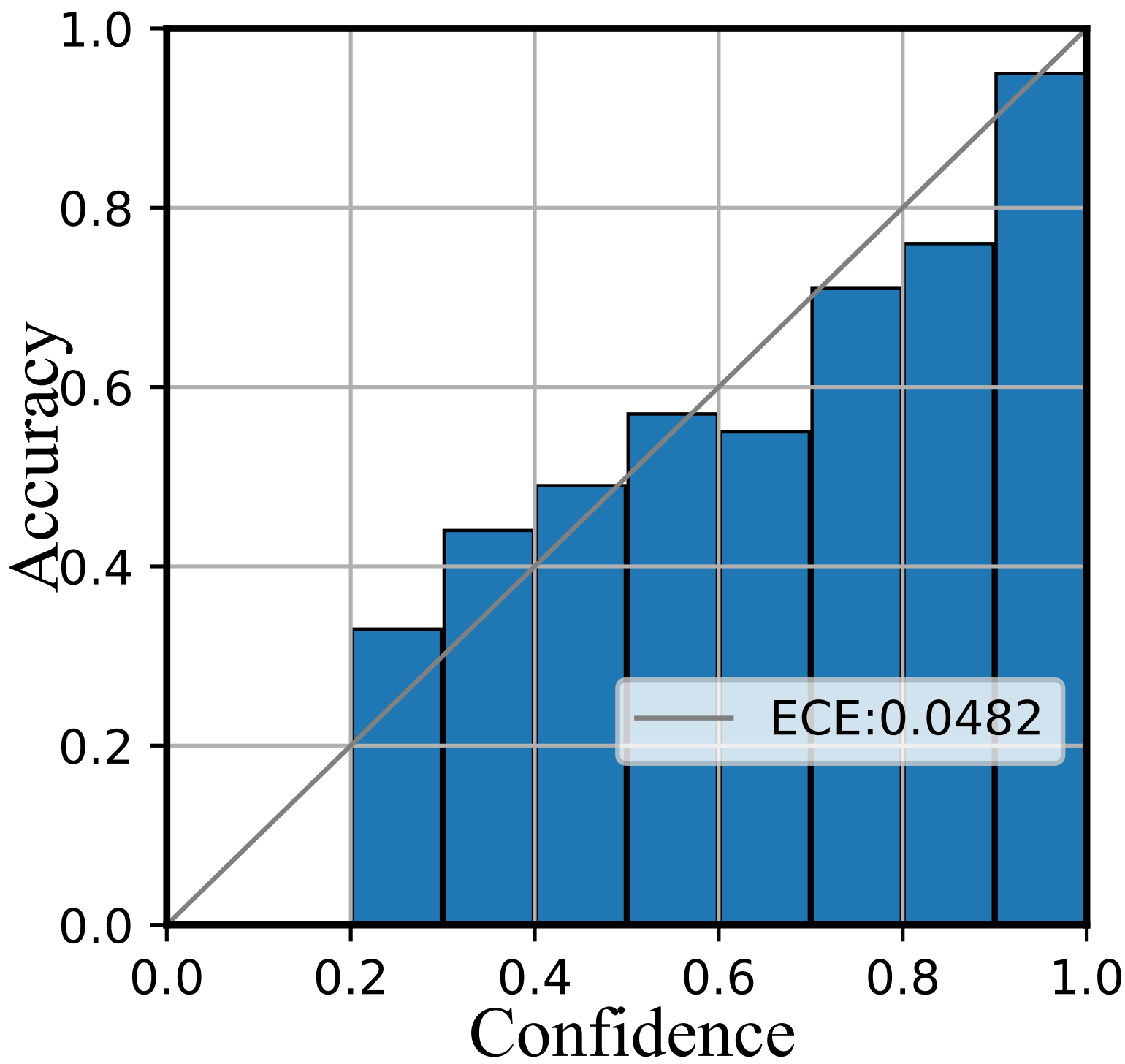}
        \caption{CoNLL2003 \name}
        \label{fig:ECE2}
    \end{subfigure}

    \begin{subfigure}[b]{0.22\textwidth}
        \includegraphics[width=\textwidth]{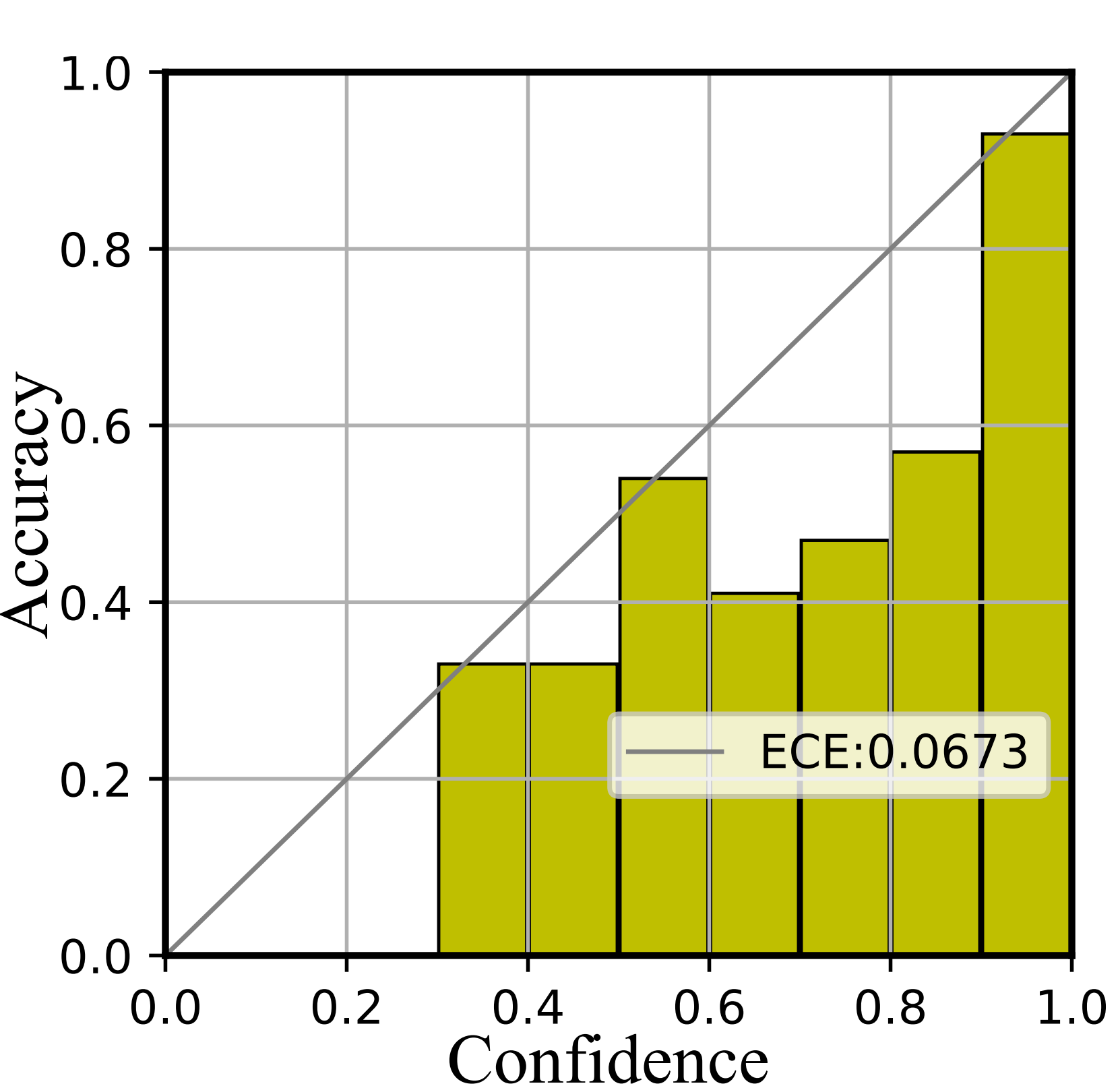}
        \caption{OntoNotes 5.0 Softmax}
        \label{fig:ECE3}
    \end{subfigure}
    \begin{subfigure}[b]{0.22\textwidth}
        \includegraphics[width=\textwidth]{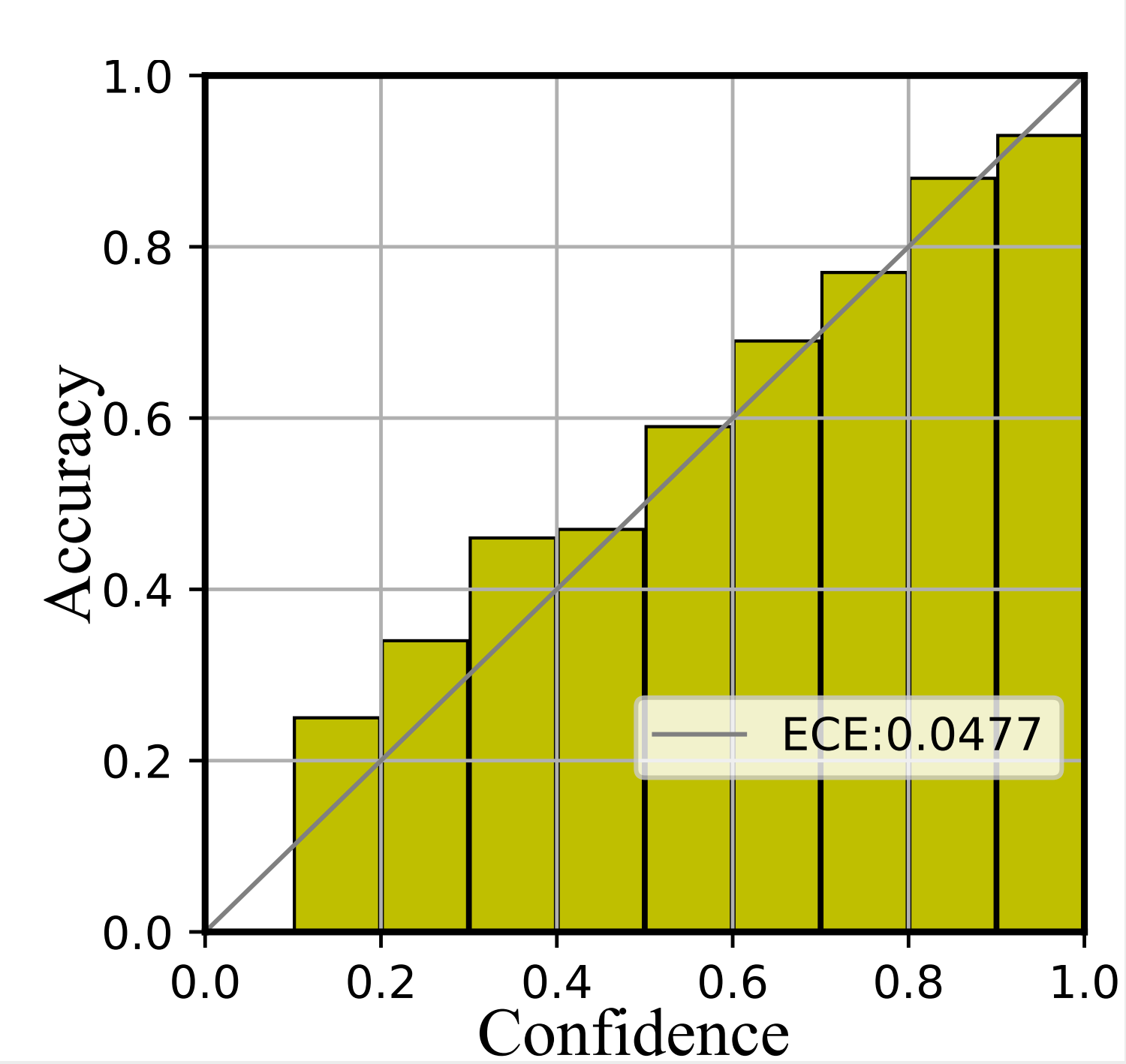}
        \caption{OntoNotes 5.0 \name}
        \label{fig:ECE4}
    \end{subfigure}
    \caption{Reliability diagrams.}
\label{fig:ECE}
\end{figure}

\begin{figure}[t!]
    \begin{subfigure}[b]{0.22\textwidth}
        \includegraphics[width=\textwidth]{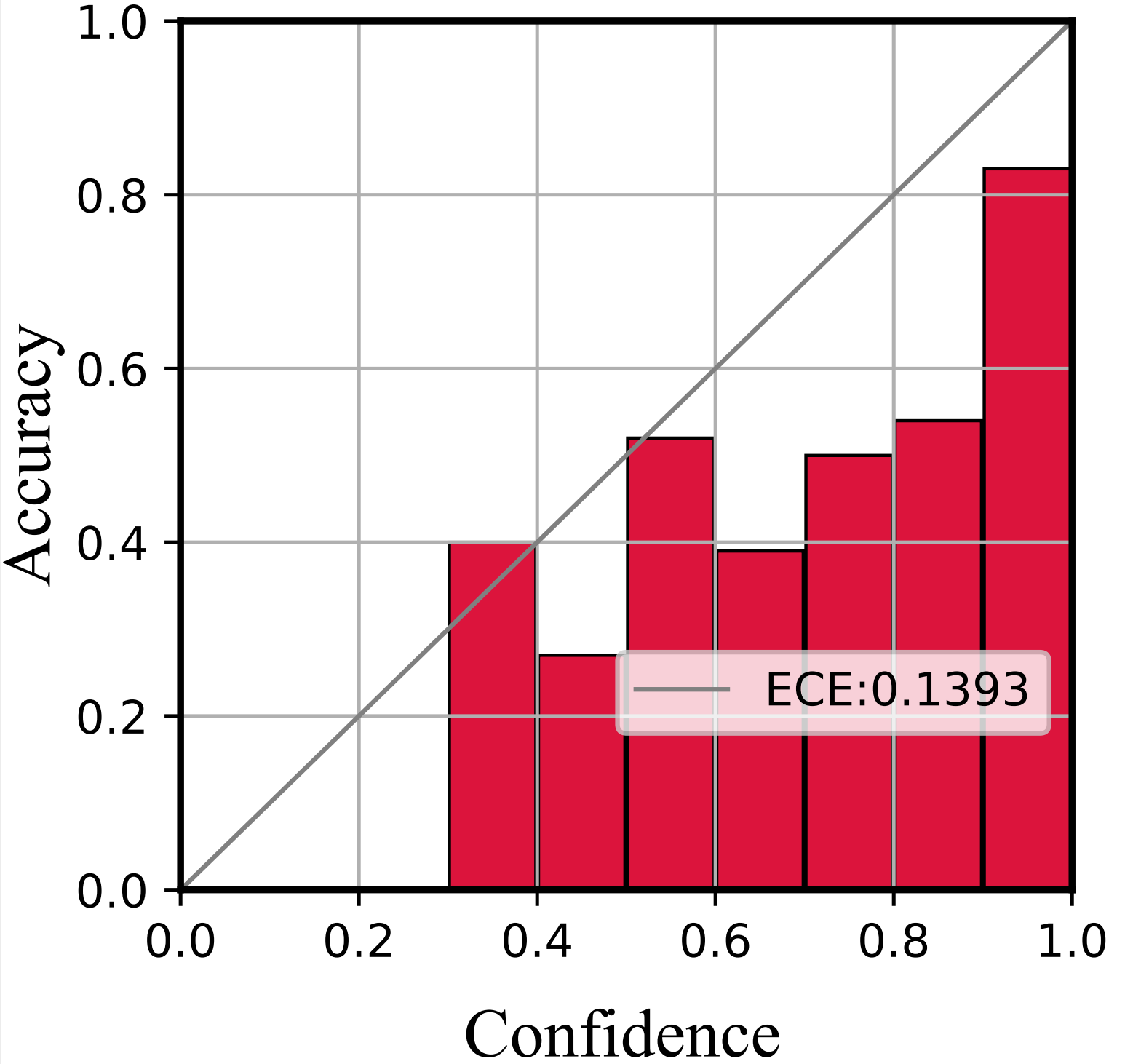}
        \caption{OOD Softmax}
        \label{fig:OOD1}
    \end{subfigure}
    \begin{subfigure}[b]{0.22\textwidth}
        \includegraphics[width=\textwidth]{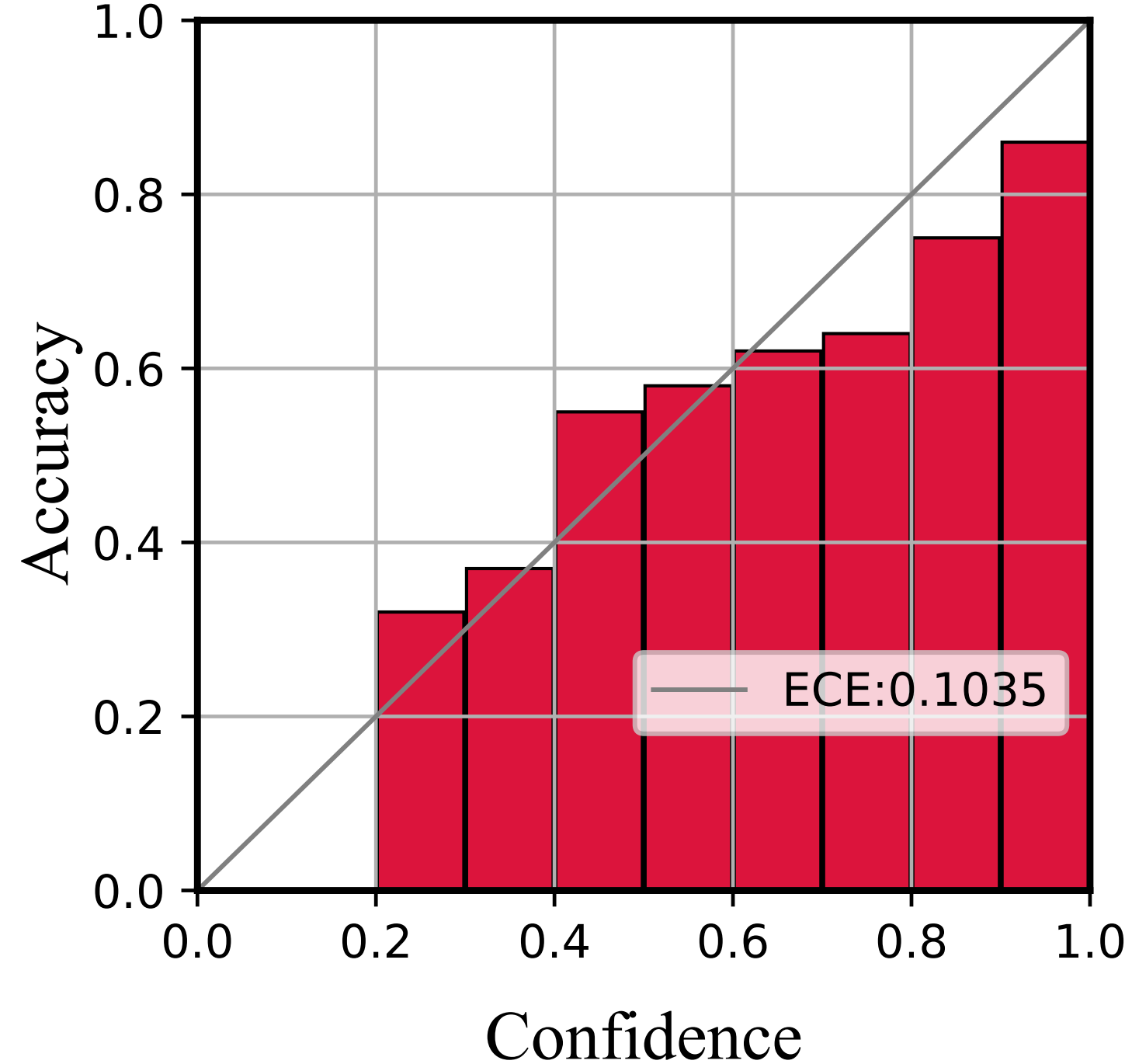}
        \caption{OOD \name}
        \label{fig:OOD2}
    \end{subfigure}
    \caption{Reliability diagrams of OOD entities. CoNLL2003 is used as the training set. The testing set of WikiGold is used for evaluating the OOD samples.}
\label{fig:OOD}
\end{figure}

\section{Additional Experimental Analysis}\label{Additional experimental_analysis}
\subsection{Reliability Diagrams}\label{relia_diagrams}

We further depict the reliability diagrams to evaluate the quality of uncertainty estimation. As shown in Figure \ref{fig:ECE} and Figure \ref{fig:OOD}, the confidence range is equally divided into ten bins. Then the subset within the same confidence range is utilized to compute the accuracy. 

As shown in Figure \ref{fig:ECE}, the confidence of Softmax represents poor accuracy, indicating it is over-confident. Then compared with Softmax, E-NER nearly approaches the perfectly calibrated line and has a much smaller ECE score. This suggests that E-NER can yield well-qualified confidence, showing it is more trustworthy. Then the observations in Figure \ref{fig:OOD} are similar, which demonstrates the reliability of the proposed approach for OOD entities.

\begin{table*}[t!]\small
    \begin{center}
        \begin{tabular}{l|l|l|l}
        \toprule  %添加表格头部粗线
        \textbf{Case} & \textbf{Sentence} & \textbf{Softmax+Entropy} & \textbf{\name}  \\
        \midrule  %添加表格中横线
        \multicolumn{1}{c|}{*}&\makecell[l]{Mapping:
        \{MIS: miscellaneous; PER: person;\\ ORG: organization; O: non-entity\}}&\multicolumn{2}{l}{\makecell[l]{Entity: \{Predcition; Confidence\%;\\ Uncertainty\%\}}}\\
        \midrule  %添加表格中横线
        \multicolumn{1}{c|}{\uppercase\expandafter{\romannumeral 1$_{ID}$}}& \multicolumn{1}{m{8.3cm}|}{A visit to the computer centre offering  $Internet^{\rm{E^1}}_{\rm{[MIS]}}$  services found a $European^{\rm{E^2}}_{\rm{[MIS]}}$ official clicking away on his mouse.} & \multicolumn{1}{m{2.7cm}|}{$\textcolor[rgb]{0.80,0.10,0.1}{{\rm{E^1}}}_{\{\rm{O}\,;\,99.9\,;\,\textcolor[rgb]{0.80,0.10,0.1}{8.0}\}}$ ${\textcolor[rgb]{0.00,0.00,1.00}{{\rm{E^2}}}}_{\{\rm{MIS}\,;\,99.9\,;\, \textcolor[rgb]{0.00,0.00,1.00}{3.0}\}}$}& \multicolumn{1}{m{2.5cm}}{$\textcolor[rgb]{0.80,0.10,0.1}{{\rm{E^1}}}_{\{\rm{O}\,;\,42.0\,;\,\textcolor[rgb]{0.80,0.10,0.1}{70.8}\}}$ ${\textcolor[rgb]{0.00,0.00,1.00}{{\rm{E^2}}}}_{\{\rm{MIS}\,;\,92.7\,;\,\textcolor[rgb]{0.0,0.0,1.0}{8.9}\}}$}\\
        \midrule  %添加表格中横线
        \multicolumn{1}{c|}{\uppercase\expandafter{\romannumeral 2$_{ID}$}}& \multicolumn{1}{m{8.3cm}|}{${Lazio^{\rm{E^1}}_{\rm{[ORG]}}}$ \ have injury doubts about striker $Pierluigi $ \ $Casiragh^{\rm{E^2}}_{\rm{[PER]}}.$} &
        \multicolumn{1}{m{2.7cm}|}{$\textcolor[rgb]{0.80,0.10,0.1}{{\rm{E^1}}}_{\{\rm{O}\,;\,98.8\,;\,\textcolor[rgb]{0.00,0.00,1.00}{7.3}\}}$ ${\textcolor[rgb]{0.00,0.00,1.00}{{\rm{E^2}}}}_{\{\rm{PER}\,;\,99.9\,;\,\textcolor[rgb]{0.00,0.00,1.00}{0.4}\}}$}& \multicolumn{1}{m{2.5cm}}{${\textcolor[rgb]{0.00,0.00,1.00}{\rm{E^1}}}_{\{\rm{ORG}\,;\,88.9\,;\,\textcolor[rgb]{0.00,0.00,1.00}{12.5}\}}$ ${\textcolor[rgb]{0.00,0.00,1.00}{{\rm{E^2}}}}_{\{\rm{PER}\,;\,98.3\,;\,\textcolor[rgb]{0.00,0.00,1.00}{2.3}\}}$}\\
        \midrule  %添加表格中横线
        \multicolumn{1}{c|}{\uppercase\expandafter{\romannumeral 3$_{OOV}$}}& \multicolumn{1}{m{8.3cm}|}{But the ${{Inthrnet}^{\rm{E^1}}_{\rm{[MIS]}}}$ , a global computer network.} & \multicolumn{1}{m{2.7cm}|}{$\textcolor[rgb]{0.80,0.10,0.1}{{\rm{E^1}}}_{\{\rm{O}\,;\,90.5\,;\,\textcolor[rgb]{0.80,0.10,0.1}{23.1}\}}$}&        \multicolumn{1}{m{2.5cm}}{$\textcolor[rgb]{0.0,0.0,1.0}{{\rm{E^1}}}_{\{\rm{MIS} \ ;  28.1\,;\,\textcolor[rgb]{0.0,0.0,1.0}{70.0}\}}$}\\
        \midrule  %添加表格中横线
        \multicolumn{1}{c|}{\uppercase\expandafter{\romannumeral 4$_{OOD}$}}& \multicolumn{1}{m{8.3cm}|}{Redesignated 65 ${Fighter\;Wing^{\rm{E^1}}_{\rm{[ORG]}}}$ on 24 July 1943.} & \multicolumn{1}{m{2.7cm}|}{$\textcolor[rgb]{0.80,0.10,0.1}{{\rm{E^1}}}_{\{\rm{O}\,;\,99.2\,;\,\textcolor[rgb]{0.80,0.10,0.1}{4.6}\}}$}&
        \multicolumn{1}{m{2.5cm}}{$\textcolor[rgb]{0.80,0.10,0.1}{{\rm{E^1}}}_{\{\rm{O}\,;\,51.3\,;\,\textcolor[rgb]{0.80,0.10,0.1}{60.7}\}}$}\\
        \bottomrule %添加表格底部粗线
    \end{tabular}
    \end{center}
    \caption{Case study of Softmax and E-NER under the span-based paradigm. The entities and their categories are already denoted in four sentences. The predicted entities with confidence (\%) and uncertainty (\%) scores are also presented. Incorrectly predicted entities are denoted by 
    ``$\textcolor[rgb]{0.80,0.10,0.1}{{\rm{Red~E}}}$'', whereas ``$\textcolor[rgb]{0.0,0.0,1.0}{{\rm{Blue~E}}}$'' represents correctly predicted entities.}
    \label{tab:study_case}  
\end{table*}

\subsection{Case Study}\label{error_case_stduy}
As presented in Table \ref{tab:study_case}, we conduct a case study by choosing four typical cases, including ID, OOV, and OOD samples. The uncertainty of Softmax is computed with entropy.

The first case contains two $\rm{MIS}$ entities. Softmax and E-NER both wrongly predict the first entity to $\rm{O}$ category, with confidence scores of 99.9\% and 42.0\%, respectively. This shows that Softmax is over-confident even for error results. Yet E-NER can output a larger uncertainty score, suggesting unsure towards the prediction. Then the second case describes two entities. Softmax wrongly predicts the first $\rm{ORG}$ entity to $\rm{O}$ with large confidence, i.e. 98.8\%. But E-NER can correctly detect the entity category as $\rm{ORG}$.

Moreover, $Inthrnet$ in the third sentence is a $\rm{MIS}$ entity, which is OOV due to misspelling. Softmax detects it as $\rm{O}$ with a confidence score of 90.5\%, showing over-confident for errors. On the contrary, E-NER assigns a large uncertainty score for the OOV sample and correctly predicts the entity category. Similarly, the last case describes an OOD entity. It can be observed that E-NER outputs a large uncertainty score compared with Softmax. 

Based on the cases and observations, we draw the following conclusions: 1) Softmax is over-confident, even for error prediction, OOV and OOD samples; 2) E-NER can recognize entities accurately and yield well-qualified uncertainties towards error, OOV and OOD samples. This contributes to the reliability and robustness of E-NER. %That's why our method can better detect OOV/OOD samples. 

\section{Implementation Details}\label{implement_details}
\subsection{Model Parameters}\label{implement_details_model}
In this paper, we implement three NER methods, including BERT-Tagger, SpanNER and Seq2Seq. The testing set is evaluated by the best model chosen by the development set. The implementation details are shown as follows.

\vspace{3pt}
\noindent
\textbf{BERT-Tagger}. BERT-Tagger\footnote{\href{https://github.com/google-research/bert}{{https://github.com/google-research/bert}}} adopts BERT-large-cased as the base encoder \cite{bert}. We set the dropout rate as 0.2, the training batch size as 16, and the weight decay as 0.02. All models in this paradigm use the Adam optimizer \cite{kingma2014adam} with a learning rate of 2e-5. Sentences are truncated to a maximum length of 256. The initial value for $\lambda_0$ is set to 1e-02.

\vspace{3pt}
\noindent
\textbf{SpanNER}. Following the original SpanNER\footnote{\href{https://github.com/neulab/spanner}{{https://github.com/neulab/spanner.}}} \cite{fu-etal-2021-spanner}, we adopt BERT-large-uncased as the base encoder \cite{bert}. The dropout rate is set to 0.2. All models in this paradigm are trained using the AdamW optimizer \cite{loshchilov2017decoupled} with a learning rate of 1e-5, with the training batch size as 10. To improve training efficiency, sentences are truncated to a maximum length of 128, and the maximum length of span enumeration is set to 4. The sampling times for MC dropout are set to 5 in the experiments. The initial value of $\lambda_0$ is set to 1e-02. We use heuristic decoding and retain the highest probability span for flattened entity recognition in span-based methods.

\vspace{3pt}
\noindent
\textbf{Seq2Seq}. Following \citet{seq2seq}, we exploit BART-Large model\footnote{\href{https://github.com/yhcc/BARTNER}{{https://github.com/yhcc/BARTNER}}}. BART model is fine-tuned with the slanted triangular learning rate warmup. The warmup step is set to 0.01. The training batch size is set to 16. The initial value of $\lambda_0$ is set to 1e-3.

\subsection{Evaluation Metrics}\label{implement_details_metrics}
\noindent \textbf{ECE}. It denotes the expected calibration error, which aims to evaluate the expected difference between model prediction confidence and accuracy \cite{Calibration}. Figure \ref{fig:OOD} depicts the difference in a geometric manner. The concrete formulation is as follows:
\begin{equation}
    \begin{aligned}
        \mathrm{ECE} = \sum_{i=1}^{|B|} \frac{N_i}{N}\lvert{\mathrm{acc}(b_i)-\mathrm{conf}(b_i)}\rvert,
    \end{aligned}
    \label{eq:ece}
\end{equation}
where $b_i$ represents the $i$-th bin and $|B|$ represents the total number of bins, setting to 10 in our experiment. $N$ denotes the number of total samples. $N_{i}$ represents the number of samples in the $i$-th bin. $\mathrm{acc}(b_i)$ denotes the accuracy and $\mathrm{conf}(b_i)$ denotes the average of confidences in the $i$-th bin.

\vspace{3pt}
\noindent \textbf{AUC}. The area under the curve (AUC)\footnote{\href{https://scikit-learn.org/stable/modules/generated/sklearn.metrics.auc.html}{{sklearn.metrics.auc.html.}}} is a commonly used metric for evaluating the performance of binary classifiers. The formulation is as follows:
\begin{equation}
    \mathrm{AUC}(f) = \frac{\sum_{t_0\in\mathcal{D}^0}\sum_{t_1\in\mathcal{D}^1}\mathbf{1}[f(t_{0})<f(t_{1})]}{|\mathcal{D}^0|\cdot|\mathcal{D}^1|}\label{auc}
\end{equation}
where $\mathcal{D}^0$ is the set of negative examples, and $\mathcal{D}^1$ is the set of positive examples. $\mathbf{1}[f(t_{0})<f(t_{1})]$ denotes an indicator function which returns 1 if $f(t_{0})<f(t_{1})$ otherwise return 0.

In this paper, we evaluate the performance of OOV/OOD detection using the AUC metric. Specifically, we consider two settings for the AUC score:

\begin{itemize}
    \item \textbf{Con}. It uses confidence as a classifier. The correct entity recognition is a positive example $\mathcal{D}^1$, and the entity recognition error is a negative example $\mathcal{D}^0$.

    \item \textbf{Unc}. It uses uncertainty as a classifier. Wrong prediction results of OOV/OOD entities are considered positive examples, denoted as $\mathcal{D}^1$. Correct prediction results of in-domain entities are considered negative examples, recorded as $\mathcal{D}^0$. These metrics assess the classifier's capability in detecting OOV/OOD entities.
\end{itemize}

\subsection{EDL Optimization Function}\label{DBM_Optimization}
In this section, we give a detailed formulation of the EDL optimization function. Eq. \ref{dir_density} introduces the density of the Dirichlet distribution. As the classification loss item of EDL, its cross-entropy loss function is as follows:
\begin{equation}
\begin{aligned}
    \mathcal{L}^{(i)}_{CLS}&\!=\!\frac{\int\left[\sum_{c=1}^{C}\!-\!y_{c}^{(i)}{\rm{log}}(p_{c}^{(i)})\right]}{B(\boldsymbol{\alpha}^{(i)})}\prod_{c=1}^C p_{c}^{\alpha_{c}^{(i)}\!-\!1}d\mathbf{p}^{(i)}\\
    &\!=\!\sum_{c=1}^C y_{c}^{(i)}\left(\psi(S^{(i)}) -\psi(\alpha_{c}^{(i)})\right).
    \label{eq:l1}
\end{aligned}
\end{equation}

The KL divergence calculation function under the Dirichlet distribution takes the following form and serves as the category penalty term in EDL:
\begin{equation}
    \begin{aligned}
        \mathcal{L}^{(i)}_{KL}&=KL[{\rm{Dir}}(\mathbf{p}^{(i)}|{\widetilde{\boldsymbol{\alpha}}^{(i)})}||{\rm{Dir}}(\mathbf{p}^{(i)}|\mathbf{1})]\\
         & = {\rm{log}}\left(\frac{\Gamma(\sum_{c=1}^C \widetilde{\alpha}_{c}^{(i)})}{\Gamma(C)\prod_{c=1}^C \Gamma(\widetilde{\alpha}_{c}^{(i)})}\right) \\ 
        &+\sum_{c=1}^C(\widetilde{\alpha}_{c}^{(i)} - 1)\left[\psi(S^{(i)}) -\psi(\sum_{j=1}^C\widetilde{\alpha}^{(i)}_{j})\right].
    \end{aligned}
    \label{eq:kl}
\end{equation}

Finally, we get the loss function for overall EDL learning:
\begin{equation}
\begin{aligned}
\mathcal{L}_{EDL}\!=\sum_{i=1}^{N}({\mathcal{L}^{(i)}_{CLS}+\mathcal{L}^{(i)}_{KL}})
    \label{eq:edl_apx}
\end{aligned}
\end{equation}
\end{document}